%% file: main.tex
\documentclass[final]{cvpr}

\usepackage{times}
\usepackage{epsfig}
\usepackage{graphicx}

\usepackage{amsmath,amssymb}
\usepackage{algorithm,listings,paralist}
\usepackage{booktabs,multirow}
\usepackage{float}

\usepackage{etoolbox}
\makeatletter
\AfterEndEnvironment{algorithm}{\let\@algcomment\relax}
\AtEndEnvironment{algorithm}{\kern2pt\hrule\relax\vskip3pt\@algcomment}
\let\@algcomment\relax
\newcommand\algcomment[1]{\def\@algcomment{\footnotesize#1}}
\renewcommand\fs@ruled{\def\@fs@cfont{\bfseries}\let\@fs@capt\floatc@ruled
  \def\@fs@pre{\hrule height.8pt depth0pt \kern2pt}%
  \def\@fs@post{}%
  \def\@fs@mid{\kern2pt\hrule\kern2pt}%
  \let\@fs@iftopcapt\iftrue}
\makeatother

\usepackage[pagebackref=true,breaklinks=true,colorlinks,citecolor=blue,linkcolor=blue,bookmarks=false]{hyperref}

\def\cvprPaperID{633} 
\def\confYear{CVPR 2021}

\begin{document}

\title{ReMix: Towards Image-to-Image Translation with Limited Data}

\author{
Jie Cao$^{1,2}$,
Luanxuan Hou$^{1,2}$,
Ming-Hsuan Yang$^{3,4,5}$,
Ran He$^{1,2}$\thanks{corresponding author},
Zhenan Sun$^{1,2}$\\
$^{1}$NLPR, CRIPAC \& CEBSIT, CASIA $^{2}$AIR, UCAS\\
$^{3}$University of California at Merced $^{4}$Google Research $^{5}$Yonsei University\\
{\tt \small \{jie.cao,\,luanxuan.hou\}@cripac.ia.ac.cn, mhyang@ucmerced.edu}\\
{\tt \small \{rhe,\,znsun\}@nlpr.ia.ac.cn}
}

\maketitle

\begin{abstract}
Image-to-image (I2I) translation methods based on generative adversarial networks (GANs) typically suffer from overfitting when limited training data is available.  
In this work, we propose a data augmentation method (ReMix) to tackle this issue. 
We interpolate training samples at the feature level and propose a novel content loss based on the perceptual relations among samples.
The generator learns to translate the in-between samples rather than memorizing the training set, and thereby forces the discriminator to generalize. 
The proposed approach effectively reduces the ambiguity of generation and renders content-preserving results. 
The ReMix method can be easily incorporated into existing GAN models with minor modifications. 
Experimental results on numerous tasks demonstrate that GAN models equipped with the ReMix method achieve significant improvements.
\end{abstract}

\section{Introduction}
In recent years, Generative Adversarial Networks (GANs)~\cite{goodfellow2014generative} have shown much progress in numerous tasks including image-to-image translation. 
Well-designed adversarial losses~\cite{goodfellow2014generative,mirza2014conditional,mao2017least,arjovsky2017wasserstein,gulrajani2017improved,mescheder2018training} provide effective domain-level supervision, making the translated results indistinguishable from the real samples. 
The GAN-based methods heavily rely on vast quantities of training examples.
For instance, Karras \etal~\cite{karras2019style,karras2020analyzing} use 70K high-quality face images to train their models. 
However, collecting a large amount of image data can be prohibitively expensive or implausible (\eg, for masterpieces by artists). 
This issue highlights the importance of training GANs with limited data.
Unfortunately, reducing the amount of training data often leads to severe model overfitting. 
Recent findings~\cite{karras2020training,zhao2020differentiable} reveal that GANs easily memorize a small training set and then render drastically degraded results in the testing set.

\begin{figure}[t]
\begin{center}
    \includegraphics[width=1.0\linewidth]{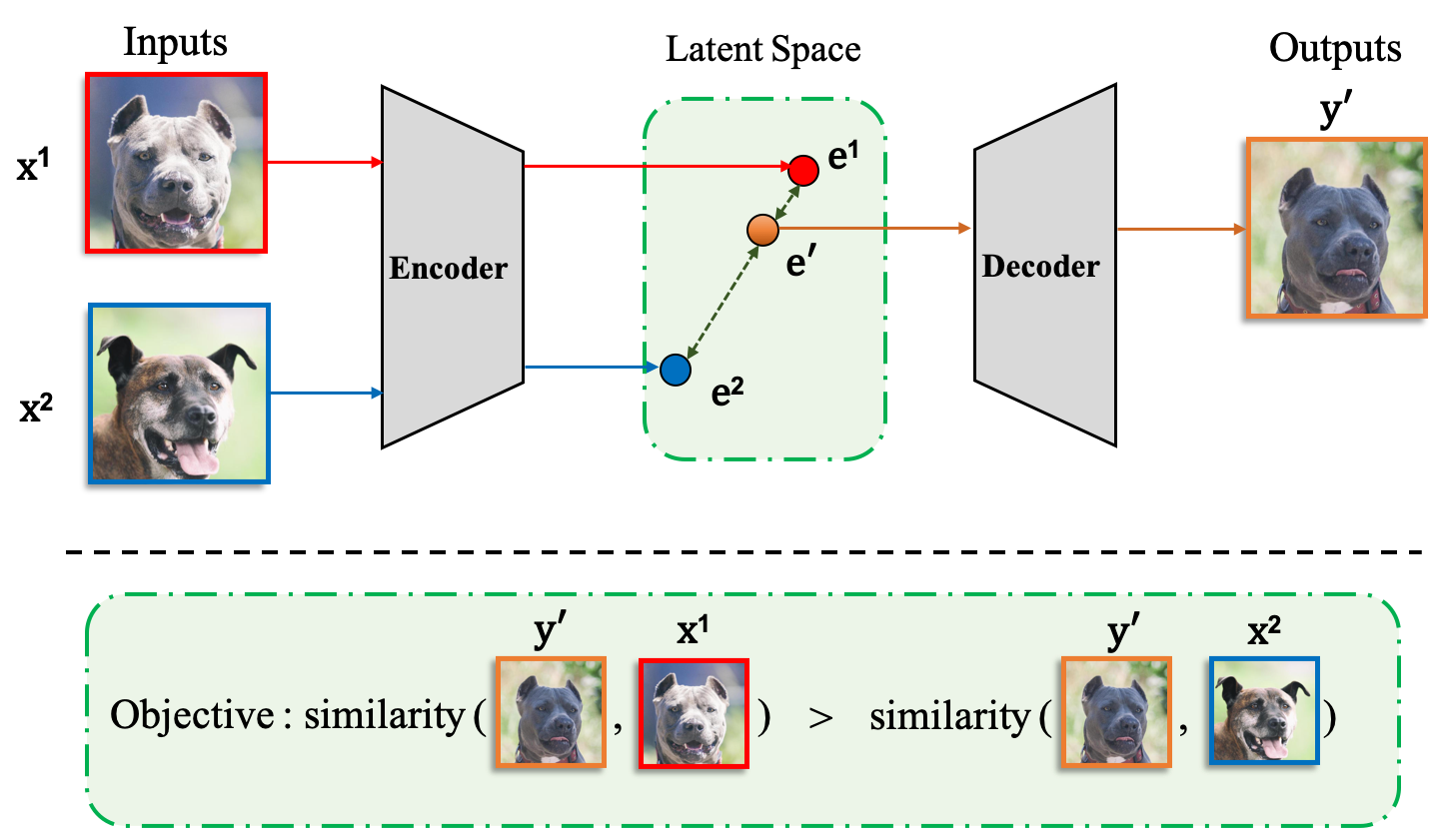}
\end{center}
\caption{
Overview of the proposed data augmentation method.
We use the image reconstruction task as an example. 
The input $\mathbf{x}$ is first encoded into representation $\mathbf{e}$ and then decoded into the output $\mathbf{y}$, and superscript indicates the index of samples. 
The interpolated data $\mathbf{e}'$ is the convex combination of $\mathbf{e}^{1}$ and $\mathbf{e}^{2}$. 
In this case, we have $d(\mathbf{e}',\mathbf{e}^{1})<d(\mathbf{e}',\mathbf{e}^{2})$, where $d$ denotes the disctance function. 
We propose to maintain $s(\mathbf{y}',\mathbf{x}^{1})>s(\mathbf{y}',\mathbf{x}^{2})$, where $s$ is the similarity measure. 
Here we omit the outputs from $\mathbf{x}^{1}$ and $\mathbf{x}^{2}$ for clarity.}
\label{fig:1}
\end{figure}

Some efforts have recently been made to tackle this problem. 
The adaption-based approaches~\cite{liu2019few,saito2020coco} use external datasets as an alternative.
They first learn a semantically related translation and then adapt it to the translation of interest.
Despite the effectiveness, these approaches require additional image collection.
Several data augmentation schemes~\cite{zhang2019consistency,karras2020training,tran2020towards,zhao2020differentiable,zhao2020image} tailored for GANs have been developed to alleviate the need for additional datasets.
They use groups of image transformations (\eg, cropping, resizing, and cutout~\cite{devries2017improved}) to augment the inputs of the discriminator.
Even with limited data, these methods can prevent the discriminator from overfitting, allowing effective adversarial supervision.
However, augmenting data for the generator is infeasible due to the problem of leaking~\cite{karras2020training}.
For the image-to-image translation tasks, these methods cannot prevent the generator from memorizing how to translate the given source images.

To facilitate training GANs with limited data in image-to-image translation, we propose a data augmentation strategy named ReMix. 
We mix source images in the feature space using convex combinations. The generator learns to map the mixed samples to the target space against overfitting. 
In addition, the discriminator is improved in the process of distinguishing the augmented fake samples. 
We present a novel content loss that maintains the perceptual relations among the samples. 
The proposed loss avoids the model from producing ambiguous results from the augmented data.
In Figure~\ref{fig:1}, the image reconstruction task is illustrated as an example.
We aim to reconstruct two samples $\mathbf{x}^{1}$ and $\mathbf{x}^{2}$, and synthesize a virtual input $\mathbf{e}'$ by interpolating the intermediate features $\mathbf{e}^{1}$ and $\mathbf{e}^{2}$. 
However, the reconstruction target for the input $\mathbf{e}'$ is unknown, so the corresponding output $\mathbf{y}'$ requires additional constraints on image content.
To this end, we propose to constrain the perceptual relationships among \{$\mathbf{x}^{1}$, $\mathbf{x}^{2}$, $\mathbf{y}'$\} based on the relationships among \{$\mathbf{e}^{1}$, $\mathbf{e}^{2}$, $\mathbf{e}'$\}.
Concretely, if $\mathbf{e}'$ is closer to $\mathbf{e}^{1}$ (or $\mathbf{e}^{2}$), we then enforce the output $\mathbf{y}'$ to be more similar to $\mathbf{x}^{1}$ (or $\mathbf{x}^{2}$) than the other one.
In this manner, we provide effective supervision and neatly sidestep estimating the targets for the interpolated inputs.

The ReMix method can be incorporated into existing methods easily. 
Only a few lines of codes are required to modify the original loss function.
In the experiments, we evaluate the proposed method on several tasks, including cross-spectrum face translation on the CASIA dataset~\cite{li2013casia}, animal face translation on the AFHQ dataset~\cite{choi2020stargan}, and image synthesis from semantic label maps on the Cityscapes dataset~\cite{cordts2016cityscapes}.
We use the state-of-the-art models~\cite{wu2018light,park2019semantic,choi2020stargan,karras2020analyzing} on these tasks as the baselines. 
Experimental results demonstrate that the models equipped with the ReMix method achieve significant improvements. 
We also train these models with 10\% available data and still get comparable performances.

The main contributions are summarized as follows:

\begin{compactitem}
   \item
   We propose a data augmentation strategy based on feature-level interpolation. 
   Our method reduces the overfitting problem of GANs, particularly for the image-to-image translation tasks.
   \item 
   We propose to maintain the perceptual relations among samples to optimize the interpolated translations.
   Our scheme reduces the ambiguity of generation and forces the model to learn content-preserving translations.
   \item 
   We achieve significant improvements in multiple image synthesis tasks.
   In addition, we produce plausible results with only 10\% training data.
\end{compactitem}

\section{Related Work}

\noindent \textbf{Unsupervised image-to-image (I2I) translation.} 
These methods aim to learn the mapping from the source domain to the target domain without paired data.
Since this problem is inherently ill-posed, the translated results will be ambiguous without additional constraints. 
To tackle this issue, existing I2I methods are constrained to preserve the image content based on pixel-level values~\cite{bousmalis2017unsupervised,shrivastava2017learning}, semantic features~\cite{taigman2016unsupervised,huang2018multimodal,lee2018diverse}, or attribute labels~\cite{bousmalis2017unsupervised}. 
The proposed loss functions, \eg, reconstruction loss and cycle consistency loss~\cite{zhu2017unpaired}, serve as the objective for content-preserving translation. 
Existing I2I methods heavily rely on large collections of high-quality images.
In this work, we propose an interpolation-based augmentation scheme for image-to-image under limited data.
To avoid ambiguous generations from the interpolated input, we develop a new loss function to preserve image content.

\noindent \textbf{Data augmentation.} Numerous methods have been developed to increase the amount of data for training deep learning models without overfitting.
Applying some content-preserving operations (\eg, flipping, rotation, and cropping) has become a routine data pre-processing step. 
To augment data for GANs, some recent approaches use adaptive~\cite{karras2020training,zhang2019consistency} or automatic~\cite{zhao2020differentiable} strategies to combine these operations. 
However, these schemes can only be applied to the discriminator and do not address the overfitting problem of the generator.

Interpolation-based augmentation methods~\cite{chawla2002smote,devries2017dataset,zhang2017mixup,beckham2019adversarial,berthelot2019mixmatch} focus on mixing training samples at the feature-level or image-level. 
Linear interpolation is simple but powerful in improving the generalization. 
For image synthesis, generating plausible interpolated results is also a desired property.
However, it remains difficult to determine supervisory signals for the interpolated inputs.
The mixup method~\cite{zhang2017mixup} assumes that the relationship between the training data and supervisory signal is linear. 
KNN interpolation algorithms~\cite{chawla2002smote,wan2017hdidx} only choose the neighbors from the same class to interpolate. 
The regularization~\cite{szegedy2016rethinking} and penalty~\cite{pereyra2017regularizing} methods can also be applied to estimate the supervisory signals.
For the image-to-image translation problems where the supervision signal is high-dimensional data, these estimations can be prone to errors. 
In contrast, our method maintains the perceptual relation among samples, which does not require the estimation of supervisory signals.

\section{Proposed Method}
\begin{figure}[t]
\begin{center}
\includegraphics[width=1.0\linewidth]{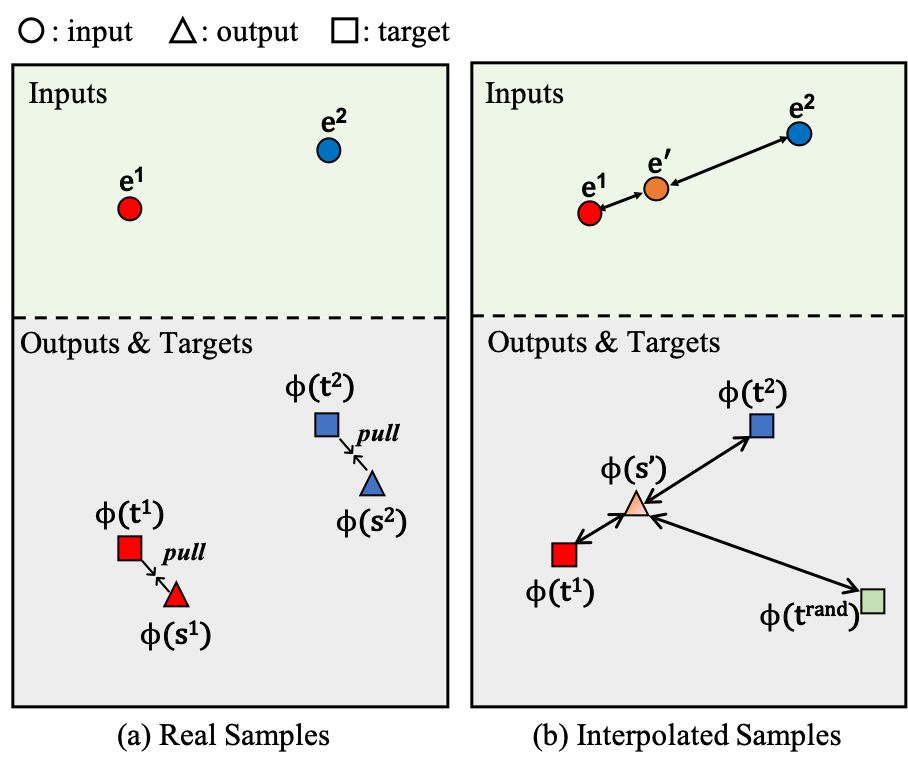}
\end{center}
   \caption{Illustration of the proposed ReMix method.
   We use colors to indicate different samples. 
   (a) For each feature $\mathbf{e}$ extracted from a real input $\mathbf{x}$, we minimize the distance between the output $\mathbf{s}$ and corresponding content target $\mathbf{t}$. 
   (b) For the interpolated feature $\mathbf{e}'= \lambda \cdot \mathbf{e}^{1} + (1 - \lambda) \cdot \mathbf{e}^{2}$, we constrain the relative similarity of the output $\mathbf{s}'$.
Concretely, if $\mathbf{e}'$ is closer/further to $\mathbf{e}^{1}$ than $\mathbf{e}^{2}$, we enforce $\phi(\mathbf{s}')$ to be closer/further to $\phi(\mathbf{t}^{1})$ than $\phi(\mathbf{t}^{2})$.
   We let $\phi$ denote the function to extract content representations.
   In addition, we enforce $\phi(\mathbf{s}')$ to be closer to $\phi(\mathbf{t}^{2})$ than $\phi(\mathbf{t}^{\text{rand}})$, and $\mathbf{t}^{\text{rand}}$ is an arbitrary sample except $\mathbf{t}^{1}$ and $\mathbf{t}^{2}$.
}
\label{fig:2}
\end{figure}

We aim to learn the mapping function from the source domain $\mathbb{X}$ to the target domain $\mathbb{Y}$.
First, we train a generator, $G: \mathbb{X} \mapsto \mathbb{Y}$, for this task.
Our goal is two-fold: 1) given $\mathbf{x} \in \mathbb{X}$, $G(\mathbf{x})$ should be indistinguishable from the samples in $\mathbb{Y}$, and 2) $G(\mathbf{x})$ should preserve certain content information. 
To this end, we optimize the adversarial loss $\mathcal{L}_{\text{gan}}$ and content loss $\mathcal{L}_{\text{con}}$. 
We formulate the objective functions for the generator $G$ and the discriminator $D$ as:
\begin{align}
\label{eq:1}
\mathcal{L}^{G}&=\sum_{(\mathbf{x},\mathbf{t}) \sim \mathbb{X},~\mathbf{y} \sim \mathbb{Y}} {\mathcal{L}_{\text{gan}}(G(\mathbf{x}),\mathbf{y})+\mathcal{L}_{\text{con}}(\phi(G(\mathbf{x})),\phi(\mathbf{t}))}, \\
\label{eq:2}
\mathcal{L}^{D}&=\sum_{\mathbf{x} \sim \mathbb{X},~\mathbf{y} \sim \mathbb{Y}} {-\mathcal{L}_{\text{gan}}(G(\mathbf{x}),\mathbf{y})},
\end{align}
where $\phi$ denotes the function to extract content representations.
The generator is trained to produce realistic samples that confuse the discriminator.
In addition, we enforce the content of $G(\mathbf{x})$ to match the content of $\mathbf{t}$, as illustrated in Figure~\ref{fig:2}(a).
Assigning $\mathbf{t}$ identical to $\mathbf{x}$ is the most common scheme for unpaired image-to-image translation, whereas other choices are also permitted by our method.
The forms of $\mathcal{L}_{\text{gan}}$ and $\mathcal{L}_{\text{con}}$ are determined according to specific tasks.
In the following, we introduce the ReMix method to augment training data for the GAN-based framework.

\subsection{Interpolation-Based Data Augmentation}
We augment the training data based on the interpolation at the feature-level. 
Let $G=G_2 \circ G_1$, where $\circ$ denotes function composition. We mix the intermediate features extracted by $G_1$. 
The interpolated data is given by:
\begin{equation}
\label{eq:3}
\mathbf{e}' = \lambda \cdot \mathbf{e}^1 + (1 - \lambda) \cdot \mathbf{e}^2,
\end{equation}
where $\mathbf{e}^1=G_1(\mathbf{x}^1)$ and $\mathbf{e}^2=G_1(\mathbf{x}^2)$. 
Here, $\mathbf{x}^1$ and $\mathbf{x}^2$ are two random samples from the source domain, and $\lambda \in [0,1]$ is the interpolation weight. 
Note that directly interpolating on the raw input $\mathbf{x}$ is a particular case in our method.

For the interpolated inputs to be useful for training, we need to translate them into content-preserving results. 
But calculating the content loss $\mathcal{L}_{\text{con}}$ for the interpolated input~$\mathbf{e}'$ requires the unknown content target $\mathbf{t}'$.
We only know $\mathbf{t}^1$ and $\mathbf{t}^2$, which are the corresponding content targets of $\mathbf{e}^1$ and $\mathbf{e}^2$, respectively.
Instead, let $\mathbf{s}'=G_2(G_1(\mathbf{e}'))$, and we constrain the perceptual relationships among \{$\mathbf{t}^1$, $\mathbf{t}^2$, $\mathbf{s}'$\} in the metric space.
Without loss of generality, we assume that $\mathbf{e}_1$ weighs more in Equation~\ref{eq:3}. We then enforce the result $\mathbf{s}'$ to satisfy the following constraint:
\begin{equation}
\label{eq:4}
\mathcal{L}_{\text{con}}\big(\phi(\mathbf{s}'),\phi(\mathbf{t}^1)\big) < \mathcal{L}_{\text{con}}\big(\phi(\mathbf{s}'),\phi(\mathbf{t}^2)\big).
\end{equation}

That is, the interpolated $\mathbf{e}'$ is closer to $\mathbf{e}_1$ than $\mathbf{e}_2$ in interpolation space, and we enforce the outputs to have an analogous relation: the corresponding output $\mathbf{s}'$ should be closer to $\mathbf{t}^1$ than $\mathbf{t}^2$ in the metric space. 
Figure~\ref{fig:2}(b) shows a visualized illustration.

Although Equation~\ref{eq:4} provides supervision to generate content-preserving results, the term $\mathcal{L}_{\text{con}}\big(\phi(\mathbf{s}'),\phi(\mathbf{t}^2)\big)$ does not have an upper bound yet. 
This means simply pushing the output $\mathbf{s}'$ away from $\mathbf{t}^{2}$ can satisfy the constraint, which is less desirable. 
Let $\mathbf{e}^{\text{rand}}=G_1(\mathbf{x}^{\text{rand}})$, where $\mathbf{x}^{\text{rand}}$ is an arbitrary sample other than $\mathbf{x}^{1}$ or $\mathbf{x}^{2}$. 
We further propose the following constraint:
\begin{equation}
\label{eq:5}
\mathcal{L}_{\text{con}}\big(\phi(\mathbf{s}'),\phi(\mathbf{t}^2)\big) < \mathcal{L}_{\text{con}}\big(\phi(\mathbf{s}'),\phi(\mathbf{t}^{\text{rand}})\big),
\end{equation}
where $\mathbf{t}^{\text{rand}}$ is the content target of $\mathbf{x}^{\text{rand}}$. 
Since $\mathbf{x}^{\text{rand}}$ does not contribute to the interpolation, $\mathbf{t}^{\text{rand}}$ should be irrelevant to the output $\mathbf{s}'$.
Therefore, we enfore $\mathbf{s}'$ to be closer to $\mathbf{t}^2$ than $\mathbf{t}^{\text{rand}}$.

We refer to the above-described scheme as ReMix for data augmentation.
We constrain the relative position of the output based on the perceptual relations among the inputs. 
Our approach provides effective supervision while allowing diverse generations. 
The translated results can be multi-modal as long as the content constraints are satisfied.

\subsection{Learning GAN Models with Limited Data}

We show how to apply the proposed ReMix method to the batch-wise training of GAN models.
For each iteration, we feed an interpolated data batch to the model with a probability of $p$. 
If the batch is not interpolated, we directly train the model with the original settings.
Otherwise, we draw two data batches, $\left\{(\mathbf{x}^{1}_{i}, \mathbf{t}^{1}_{i})\right\}_{i=1}^{n}$  and $\left\{(\mathbf{x}^{2}_{i}, \mathbf{t}^{2}_{i})\right\}_{i=1}^{n}$ , where $n$ denotes the batch size.
Similar to the mixup method~\cite{zhang2017mixup}, we calculate the interpolation weight by:
\begin{align}
\label{eq:6}
\mu &= \text{Beta}(\alpha, \alpha), \\
\label{eq:7}
\lambda &= \text{max}(\mu, 1 - \mu),
\end{align}
where $\text{Beta}(\alpha, \alpha)$ denotes the beta distribution parameterized by $\alpha$. 
We then obtain the augmented inputs $\left\{\mathbf{e}'_{i}\right\}_{i=1}^{n}$ by the interpolation scheme formulated in Equation~\ref{eq:3}.
Note that $\mathbf{e}^{1}_{i}$ always weighs more in the interpolation because we have $\lambda \geq 0.5$.

We compute the adversarial loss $\mathcal{L}_{\text{gan}}$ using the augmented batch for domain-level supervision. 
For the content supervision described in Equations~\ref{eq:4} and \ref{eq:5}, we have:
\begin{align}
\label{eq:8}
\mathcal{L}_{p}&=\sum_{i} \text{max}\big\{0,~\mathcal{L}_{\text{con}}\big(\phi(\mathbf{s}'_{i}),\phi(\mathbf{t}^{1}_{i})\big) - \mathcal{L}_{\text{con}}\big(\phi(\mathbf{s}'_i),\phi(\mathbf{t}^{2}_{i})\big)\big\}, \\ 
\label{eq:9}
\mathcal{L}_{n}&=\sum_{i} \text{max}\big\{0,~\mathcal{L}_{\text{con}}\big(\phi(\mathbf{s}'_{i}),\phi(\mathbf{t}^{2}_{i})\big) - \bar{a}\}.
\end{align}

We minimize $\mathcal{L}'_{con}=\mathcal{L}_{p} + \mathcal{L}_{n}$ for the interpolated inputs, which is referred to as the relative form of $\mathcal{L}_{con}$. 
We initialize $\bar{a}$ to 0 and update it dynamically during training. 
Concretely, we first compute:
\begin{equation}
\label{eq:10}
a = \sum_{i} \mathcal{L}_{\text{con}}\big(\phi(\mathbf{s}'_{i}),\phi(\mathbf{t}^{2}_{j})\big),
\end{equation}
where $j \neq i$, so $\mathbf{e}^{2}_{j}$ does not contribute to the interpolation of $\mathbf{e}'_{i}$. 
Hence, $a$ denotes the mean distance of the unrelated output-target pairs within the training batch. 
Then, we adopt a momentum update of $\bar{a}$:
\begin{equation}
\label{eq:11}
\bar{a} \leftarrow m \cdot \bar{a} + (1-m) \cdot (a - \bar{a}),
\end{equation}
where we set the momentum coefficient $m$ to 0.99. 
Algorithm~\ref{alg:1} shows the main step to train the generator with the ReMix data augmentation method. 
For the discriminator, the process is similar. 
We compute one single content loss in the relative form, whereas the ReMix method can also be applied to the case with multiple content losses. 
Each loss can be calculated in the relative form independently. 

\subsection{Comparison with Existing Methods}

In contrast to existing approaches, the ReMix method does not rely on estimating the corresponding target $\mathbf{t}'$ for each interpolated input $\mathbf{e}'$.
For example, the scheme by Zhang~\etal~\cite{zhang2017mixup} assumes the relationship between the training data and supervision signal is linear. 
Hence, given the interpolation weight $\lambda$ for the inputs, this scheme~\cite{zhang2017mixup} computes:
\begin{equation}
\mathbf{t}' = \lambda \cdot \mathbf{t}^1 + (1 - \lambda) \cdot \mathbf{t}^2.
\end{equation}

In addition, this method proposes to directly use the supervisory signal of the sample that weighs more with:
\begin{equation}
\label{eq:WM}
\mathbf{t}'=
\begin{cases}
\mathbf{t}^1, & \text{if } \lambda \geq 0.5, \\
\mathbf{t}^2, & \text{otherwise.}
\end{cases}
\end{equation}

Furthermore, regularizations can be used in the estimation the content target.
For instance, based on the LSR method~\cite{szegedy2016rethinking}, we can clamp the weight $\lambda$ into a predefined range $[\lambda_{\text{min}},\lambda_{\text{max}}]$ to interpolate the content target.
Other tricks like noise injection, nearest-neighbor interpolation~\cite{chawla2002smote,wan2017hdidx} can also be used.

The approaches described above use the estimated input-target pairs to augment the training data. 
For the classification tasks where the target $\mathbf{t}'$ is a label, they are shown to be effective. 
However, in the image-to-image translation tasks, we use raw images or high-dimensional features as supervision signals, which are substantially more difficult to estimate.
Inaccurate estimations may negatively affect the quality of the augmented training data. 
We evaluate the ReMix method against these schemes for multiple image-to-image translation tasks.

\begin{algorithm}[t]
\caption{Pseudocode of the ReMix method.}
\label{alg:1}
\definecolor{codeblue}{rgb}{0.25,0.5,0.5}
\lstset{
  backgroundcolor=\color{white},
  basicstyle=\fontsize{7.2pt}{7.2pt}\ttfamily\selectfont,
  columns=fullflexible,
  breaklines=true,
  captionpos=b,
  commentstyle=\fontsize{7.2pt}{7.2pt}\color{codeblue},
  keywordstyle=\fontsize{7.2pt}{7.2pt},
}
\begin{lstlisting}[language=python]
# D : Discriminator, (N * C * H * W) -> N
# G: Generator, which consists of G1 and G2
# G1: (N * C * H * W) -> (N * C' * H' * W')
# G2: (N * C' * H' * W') -> (N * C * H * W)
# gan : the adversarial loss, N -> N
# phi: extracting content, (N * C * H * W) -> (N * E)
# con: the content loss, (N * E) -> N

for batch1, batch2 in data_loader:
    # the probability of augmentation is p
    if p > rand(0, 1):
        # x, t : input and target, (N * C * H * W)
        x1, t1 = batch1 
        x2, t2 = batch2
        e1, e2 = G1.forward(x1), G1.forward(x2)
        
        # interpolating the input
        mu = beta.draw() # the beta distribution
        lambda = max(mu, 1-mu)
        e_prime = lambda * e1 + (1 - lambda) * e2
        
        # calculating the adversarial loss
        s_prime = G2.forward(e_prime)
        prediction = D.forward(s_prime)
        loss_gan = gan(prediction).mean()
        
        # calculating the relative content loss
        d1 = con(phi(s_prime), phi(t1))
        d2 = con(phi(s_prime), phi(t2))
        # clamp : clamp all elements into [0, Infinity]
        l_p = clamp(d1 - d2).mean()
        l_n = clamp(d2 - a_mean).mean()
        loss_con = l_p + l_n
        
        # update of Generator
        loss = loss_gan + loss_con
        loss.backward()
        update(G.parameters)
    
        # momentum update of a_mean
        # shuf : shuffle data along the batch axis
        a = con(phi(s_prime), phi(shuf(t2))).mean()
        a_mean = m * a_mean + (1 - m) * (a - a_mean)
\end{lstlisting}
\end{algorithm}

\section{Experiments and Analysis}

\input{afhq1}

We consider three practical tasks, \ie, NIR-to-VIS face translation, animal face translation, and image synthesis from semantic label maps. 
We first introduce the datasets and implementation details.

The Animal Faces-HQ dataset (AFHQ)~\cite{choi2020stargan} provides animal faces of three domains: cat, dog, and wildlife. 
Each category contains about 5,000 images. 
We aim to train a single model to learn the translations among these domains.
The StarGAN v2~\cite{choi2020stargan} is used as the baseline for this task. 
We interpolate the output of the style encoder in the baseline model~\cite{choi2020stargan}.
In our ReMix method, we modify the style reconstruction loss to the relative form.

The CASIA NIR-VIS 2.0 Face Dataset~\cite{li2013casia} contains near-infrared (NIR) and visible (VIS) images of 725 subjects. 
There are large variations of the same identity, including lighting, expression, pose, and accessories. 
For the NIR-to-VIS face translation, we use the LightCNN-29v2~\cite{wu2018light} and StyleGAN2~\cite{karras2020analyzing} to build an encoder-decoder network. 
The LightCNN~\footnote{\url{https://github.com/AlfredXiangWu/LightCNN}} is pre-trained, and we choose to interpolate its outputs. 
We train the StyleGAN from scratch using the default settings~\cite{karras2020analyzing}.
In addition, we add the L1 distance loss~\cite{isola2017image} in the pixel space as the content supervision. 
When learning GAN models with the interpolated data, we use the relative form of the L1 distance loss.

The Cityscapes dataset~\cite{cordts2016cityscapes} contains 3,500 street scene images and the corresponding semantic label maps. 
We use the SPADE Net~\cite{park2019semantic} for translating the label maps to scenes. 
We directly interpolate the raw inputs in this task.
The baseline model uses the perceptual loss~\cite{johnson2016perceptual} guided by VGGNet~\cite{simonyan2014very}. 
For ReMix, we modify this loss to the relative form.

We only modify the mentioned losses for the ReMix method, and the other losses remain the same.
We set the probability of augmentation to 0.25 for each iteration. 
Similar to the mixup method~\cite{zhang2017mixup}, we set the hyper-parameter $\alpha=0.2$ for the beta distribution.

We implement these baselines using the released source codes.
The input resolution is $512 \times 256$ on the Cityscapes dataset~\cite{cordts2016cityscapes} and $256 \times 256$ for the others. 
We change the dimension of the input latent in StyleGAN2~\cite{karras2020analyzing} to $256$.
Except for this point, we do not make any modifications to the network architectures. 
We use the recommended training settings in the original work for each baseline model, including the batch size, optimizer, training iterations, and loss weights.
To determine the value of the augmentation probability in our ReMix method, we conduct a grid search on the AFHQ dataset and use the FID score as the metric.
We use the found value for all the experiments without hyper-parameter tuning.

\subsection{Animal Face Translation}

\input{tab1}
\input{afhq2}
\input{tab2}
\input{casia1}

The first task is to change the species of the given animal face. 
If a reference image is available, an encoder extracts the style representation from it.
Then, the generator mixes the style with the content of the input, producing the translated result.
Otherwise, given a one-hot class label, the generator draws a latent code from a prior distribution as the style representation.
The two types of tasks are referred to as reference-guided translation and latent-guided translation, respectively.

We randomly choose 500 images for each class, which is about 10\% of the full training set.
Then, we train the models under the 10\% data settings.
We evaluate our method against existing interpolation approaches, including the mixup~\cite{zhang2017mixup} and WM (Equation~\ref{eq:WM}) schemes.
Figure~\ref{fig:afhq1} shows some synthesized images by the evaluated methods.
The baseline model suffers from the overfitting problem and generates some unrealistic texture details.
Overall, the proposed method synthesizes images with higher visual quality than other schemes.
Figure~\ref{fig:afhq2} shows diverse translation results by our method. 
Given a source image, we generate diverse results by randomly sampling multiple reference images.
These results show that our approach can generate distinctive styles while preserving content information.

We also evaluate the quality of synthesized images using Fr\'echet Inception Distance (FID)~\cite{heusel2017gans} and Learned Perceptual Image Patch Similarity (LPIPS)~\cite{zhang2018unreasonable}. 
The FID metric~\cite{heusel2017gans} measures the Wasserstein distance between two image sets.
We extract the features from the last average pooling layer of the Inception-v3 model~\cite{szegedy2016rethinking} to calculate the FID score.
The LPIPS score \cite{zhang2018unreasonable} measures the diversity of images using the L1 distance in the feature space, and the pre-trained AlexNet \cite{krizhevsky2012imagenet} is used as the feature extractor.
We compute the FID and LPIPS scores for every pair of the image domains (\eg, dog$\rightarrow$cat, cat$\rightarrow$wildlife) and report the average values.

Table~\ref{tab:1} shows the FID and LPIPS scores. 
We evaluate the methods under both the 10\% and 100\% data settings.
Our approach performs favorably against existing augmentation methods in terms of these quantitative metrics. 
The FID scores indicate that our results are more similar to the real data.
The LPIPS score of our method with 10\% data is higher than that of the baseline trained with the full training set.
These results demonstrate the proposed method is effective for diverse and realistic image translation.

\subsection{Cross-Spectrum Face Translation}

\input{tab3}
\input{city1}
The second task is to translate the input NIR face into the VIS domain and preserve the identity (content) information. 
Prior works evaluate image generation methods based on the ``recognition via generation'' protocol~\cite{he2019adversarial}.
That is, given a NIR face image, we use the translated result for recognition.
Using the same protocol, we use the 357 identities in the training set of the first fold~\footnote{There are 10 fold experiments on the CASIA NIR-VIS 2.0 Face Dataset. Image generation methods are usually evaluated on the first fold.} to train our model.
The remaining images are used for testing.
Table~\ref{tab:2} shows Rank-1 accuracy and the verification rates of different methods. 
The competing data augmentation methods are the mixup~\cite{zhang2017mixup} and WM (Equation~\ref{eq:WM}) methods.
Our method performs favorably against the other schemes in terms of both rank accuracy and verification rates.
These experimental results show that our method reduces the domain gap between the NIR and VIS face images effectively.

We also consider an extreme scenario where only one NIR-VIS image pair of each identity is used for training. 
That is, the training set consists of only $357$ pairs of images.
We show the generated samples from the testing set in Figure~\ref{fig:casia1}.
We observe that the ReMix method synthesizes plausible results even with limited data. 
The baseline model and WM cannot produce satisfactory results due to model overfitting.
The appearances synthesized by the mixup method are realistic, but the identities look different from the input NIR images.

\subsection{Image Synthesis from Semantic Label Maps}
Given a semantic layout, we train the translation models to synthesize a photorealistic image.
The official training split of the Cityscapes dataset~\cite{cordts2016cityscapes} consists of 3000 pairs of image and semantic label maps.
We train the models under the 10\% and 100\% data settings for this task.
We use FID to measure the distance between the distributions of the real images and the distribution of the synthesized results.
In addition, we perform semantic segmentation on the synthesized images and then evaluate how well the predicted results match the input label maps.
Similar to prior work~\cite{park2019semantic}, we use DRN-D-105~\cite{yu2017dilated} to measure the segmentation accuracy. 
Table~\ref{tab:3} reports the FID scores and the predicted segmentation accuracy of different methods.
In Figures~\ref{fig:city1}, we provide samples of the translation results under the 10\% data setting.
The competing methods in the table are also the mixup and WM methods.
We observe that the ReMix method performs favorably against the state-of-the-art methods in terms of the quantitative metrics.
Our method produces results with better visual quality and fewer artifacts.
In contrast, the performances under the 10\% data setting degrade significantly for the other approaches.

\section{Conclusion}
We introduce an interpolation-based data augmentation method to tackle the overfitting problem of GANs.
In addition, we present to maintain the perceptual similarity among samples to reduce the ambiguity of generation.
The proposed approach renders content-preserving results from the interpolated inputs, facilitating the model training in image-to-image translation.
We demonstrate that our method vastly improves the image quality and quantitative metrics in numerous tasks, especially when the training data is limited.

\section{Acknowledgement}

This work is funded by National Natural Science Foundation of China (Grant~No.~U1836217) and Beijing Natural Science Foundation (Grant~No.~JQ18017). M.-H. Yang is partially supported by NSF CAREER 1149783.

\clearpage

{\small

\input{main.bbl}
}

\end{document}

%% file: afhq1.tex
\begin{figure*}[t]
\begin{center}
\includegraphics[width=1.0\linewidth]{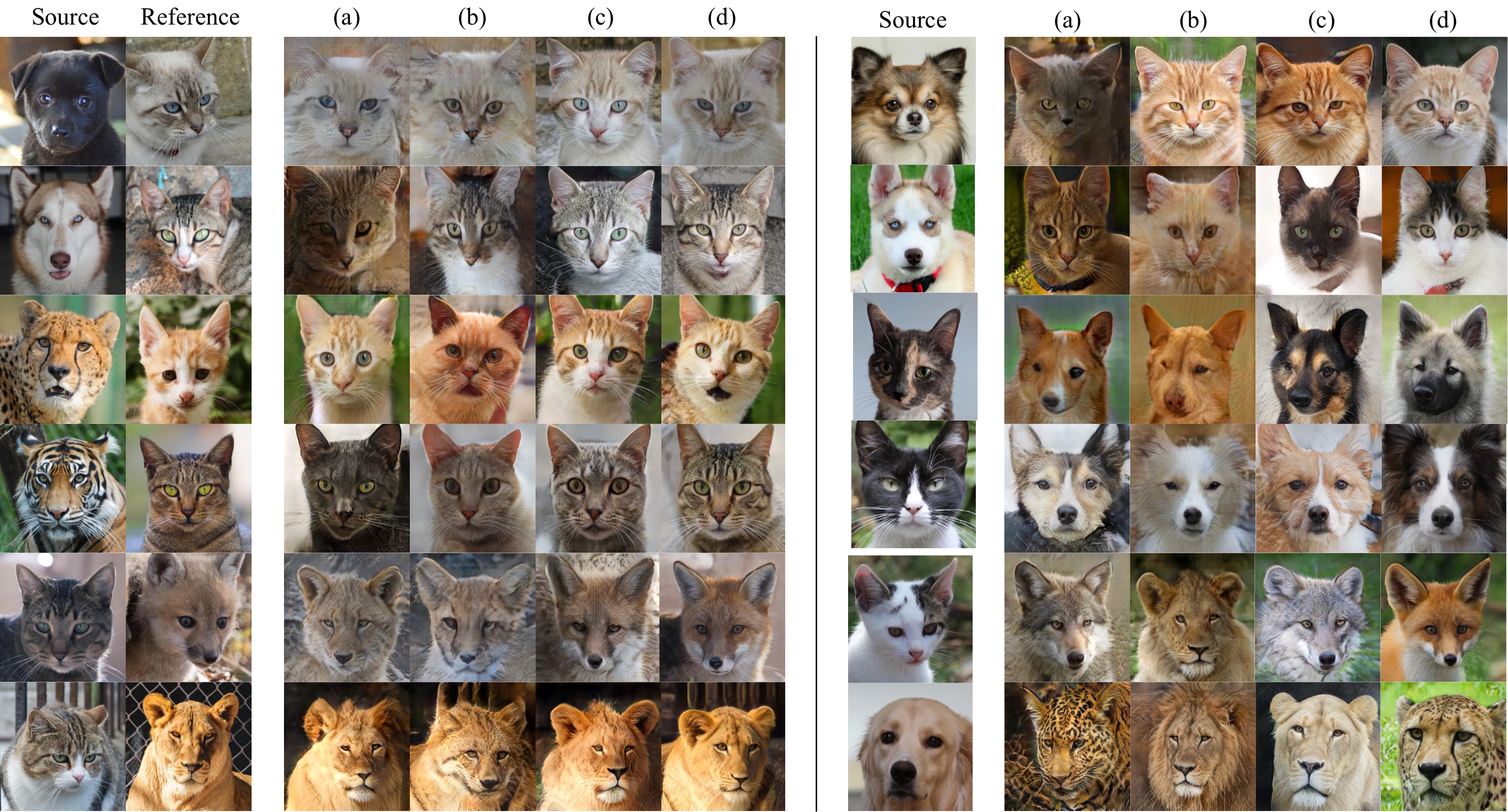}
\end{center}
\vspace{1mm}
\caption{
Visual examples synthesized by different methods with 10 \% training data on the AFHQ dataset~\cite{choi2020stargan}. 
The left part is the results of reference-guided translation, and the right part is the results of latent-guided translation.
The column of results are (a) StarGAN v2~\cite{choi2018stargan} (baseline), (b) baseline + WM~(Equation \ref{eq:WM}), (c) baseline + mixup~\cite{zhang2017mixup}, and (d) baseline + ReMix (ours).
}
\label{fig:afhq1}
\end{figure*}

%% file: tab1.tex
\begin{table*}[t]
\begin{center}
\small
\caption{
Fr\'echet Inception Distance (FID, lower is better) and Learned Perceptual Image Patch Similarity (LPIPS, higher is better) of different methods on the AFHQ dataset~\cite{choi2020stargan}.
The WM method is described in Equation~\ref{eq:WM}.
}
\vspace{4mm}
\label{tab:1}%
    \begin{tabular}{lccccccccccc}
    \toprule
    \multirow{3}[6]{*}{Method} & \multicolumn{5}{c}{Latent-guided translaion} &       & \multicolumn{5}{c}{Reference-guided translation} \\
\cmidrule{2-6}\cmidrule{8-12}          & \multicolumn{2}{c}{100\% data} &       & \multicolumn{2}{c}{10\% data} &       & \multicolumn{2}{c}{100\% data} &       & \multicolumn{2}{c}{10\% data} \\
\cmidrule{2-3}\cmidrule{5-6}\cmidrule{8-9}\cmidrule{11-12}          &~FID$\downarrow$~&~LPIPS$\uparrow$~& & FID$\downarrow$~&LPIPS$\uparrow$~&~& FID$\downarrow$~&~LPIPS$\uparrow$~& &~FID$\downarrow$~&~LPIPS$\uparrow$~\\
    \midrule
    Baseline : StarGAN v2~\cite{choi2020stargan} & 16.18  & 0.450 & & 46.02 & 0.431 & & 19.78 & 0.432  &  & 38.42 & 0.402 \\
    Baseline \textbf{+} WM & 20.03 & 0.484 & & 41.36 & \textbf{0.477} & & 23.64 & 0.475 & & 45.88 & 0.455 \\
    Baseline \textbf{+} mixup~\cite{zhang2017mixup} & 15.91 & 0.453 & & 28.15 & 0.466 &  & 18.67 & 0.453 & & 27.34 & 0.451  \\
    \midrule
    Baseline \textbf{+} ReMix (ours) & \textbf{15.22} & \textbf{0.491} & & \textbf{21.82} & 0.471 &       & \textbf{15.56} & \textbf{0.481} &  & \textbf{22.92} & \textbf{0.460} \\
    \bottomrule
    \end{tabular}%
\end{center}
\end{table*}%

%% file: afhq2.tex
\begin{figure*}[t]
\begin{center}
\includegraphics[width=1.0\linewidth]{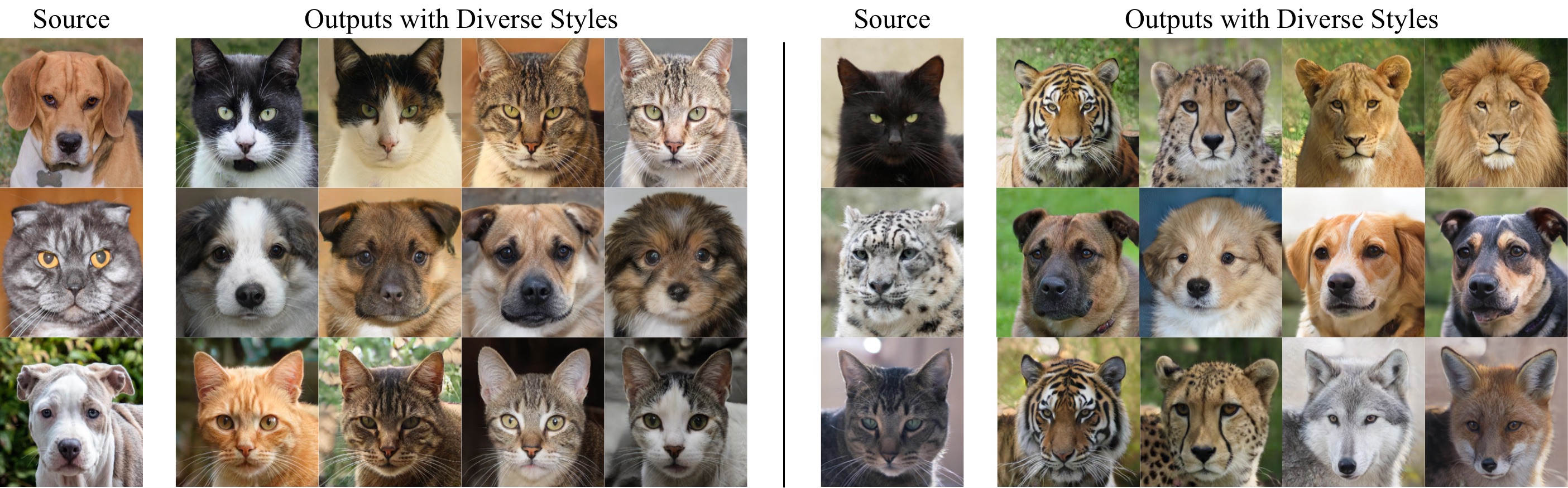}
\end{center}
\vspace{1mm}
\caption{
Diverse translation results on the AFHQ dataset~\cite{choi2020stargan}. 
Our model can learn to generate diverse high-quality results using only 10\% data in the training set.
}
\label{fig:afhq2}
\end{figure*}

%% file: tab2.tex
\begin{table*}[t]
\begin{center}
\small
\caption{
Rank-1 accuracy (\%) and verification rate (\%, VR) of different methods on the CASIA NIR-VIS 2.0 dataset (the first fold). 
FAR denotes the false acceptance rate.
The performances are evaluated according to the ``recognition via generation'' protocol~\cite{he2019adversarial}.
We use LightCNN-29v2~\cite{wu2018light} and StyleGAN2~\cite{karras2020analyzing} to build an encoder-decoder network as the baseline.
The WM method is described in Equation~\ref{eq:WM}.
``raw input'' means we directly use the LightCNN model to match the NIR faces with the VIS faces.
}
\vspace{4mm}
\label{tab:2}%
    \begin{tabular}{lccc}
    \toprule
    Method &~~Rank-1~~&~~VR@FAR=1\%~~&~~VR@FAR=0.1\%~~\\
    \midrule
    Raw Input & 96.84 & 99.10 & 94.68 \\
    Baseline~\cite{wu2018light,karras2019style} & 93.13 & 94.22 & 88.79 \\
    Baseline \textbf{+} WM & 91.32 & 92.57 & 81.27 \\
    Baseline \textbf{+} mixup~\cite{zhang2017mixup} & 97.66 & 99.38 & 97.59 \\
    \midrule
    Baseline \textbf{+} ReMix (ours) & \textbf{98.18} & \textbf{99.63} & \textbf{98.11} \\
    \bottomrule
    \end{tabular}%
\end{center}
\end{table*}%

%% file: casia1.tex
\begin{figure*}[t]
\begin{center}
\includegraphics[width=1.0\linewidth]{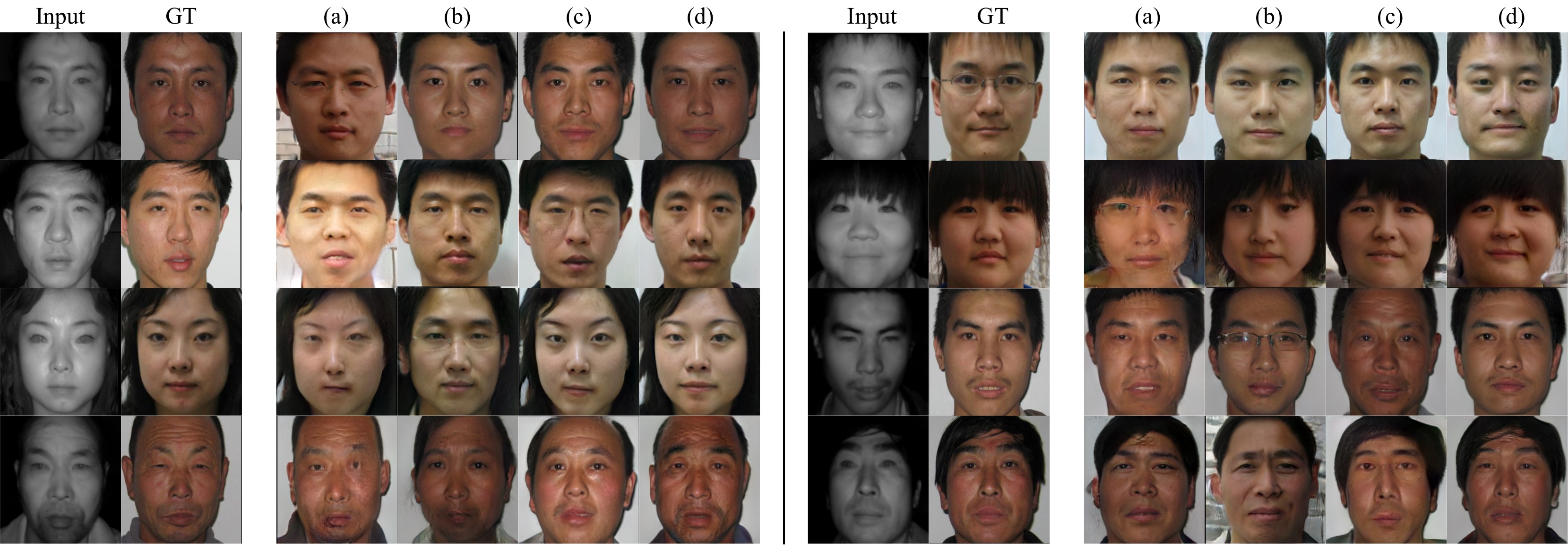}
\end{center}
\vspace{1mm}
\caption{
Face cross-spectrum translation results from the testing set of the CASIA NIR-VIS 2.0 dataset~\cite{he2019adversarial}.
We train models with only 357 pairs of NIR-VIS images.
The column of results are (a) the baseline (LightCNN~\cite{wu2018light} + StyleGAN~\cite{karras2020analyzing}), (b) baseline + WM~(Equation \ref{eq:WM}), (c) baseline + mixup~\cite{zhang2017mixup}, and (d) baseline + ReMix (ours). 
``GT'' denotes the ground-truth VIS image (not strictly paired).
}
\label{fig:casia1}
\end{figure*}

%% file: tab3.tex
\begin{table*}[t]
\small
\caption{
Semantic segmentation scores (higher is better) and Fr\'echet Inception Distance (FID, lower is better) of different methods on the Cityscapes dataset~\cite{cordts2016cityscapes}.
``mIou'' denotes the mean Intersection-Over-Union metric, and ``accu'' denotes the pixel-wise accuracy.
``real data'' denotes the results evaluated on the real images, which is the theoretical upper bound we can achieve.
}
\vspace{4mm}
\label{tab:3}%
\begin{center}
    \begin{tabular}{lccccccc}
    \toprule
          & \multicolumn{3}{c}{100\% training data} &       & \multicolumn{3}{c}{10\% training data} \\
\cmidrule{2-4}\cmidrule{6-8}    Method &~~~mIoU$\uparrow$~~~&~~~accu$\uparrow$~~~&~~~FID$\downarrow$~~~& &~~~mIoU$\uparrow$~~~&~~~accu$\uparrow$~~~&~~~FID$\downarrow$~~~\\
    \midrule
    Real Data  & 75.6 & 84.8 & - & & - & - & - \\
    Baseline: SPADE Net~\cite{park2019semantic} & 62.3 & 81.9 & 71.8 & & 48.3 & 68.2 & 85.9 \\
    Baseline \textbf{+} WM  & 51.1 & 80.2 & 95.5 & & 45.4 & 57.3 & 108.8 \\
    Baseline \textbf{+} mixup~\cite{zhang2017mixup} & 65.5 & 82.3 & 64.7 & & 59.7 & 72.1 & 71.5 \\
    \midrule
    Baseline \textbf{+} ReMix (ours)  & \textbf{70.3} & \textbf{82.7} & \textbf{50.1} &       & \textbf{62.1} & \textbf{74.4} & \textbf{68.0} \\ 
    \bottomrule
    \end{tabular}%
\end{center}
\end{table*}%

%% file: city1.tex
\begin{figure*}[t]
\begin{center}
\includegraphics[width=1.0\linewidth]{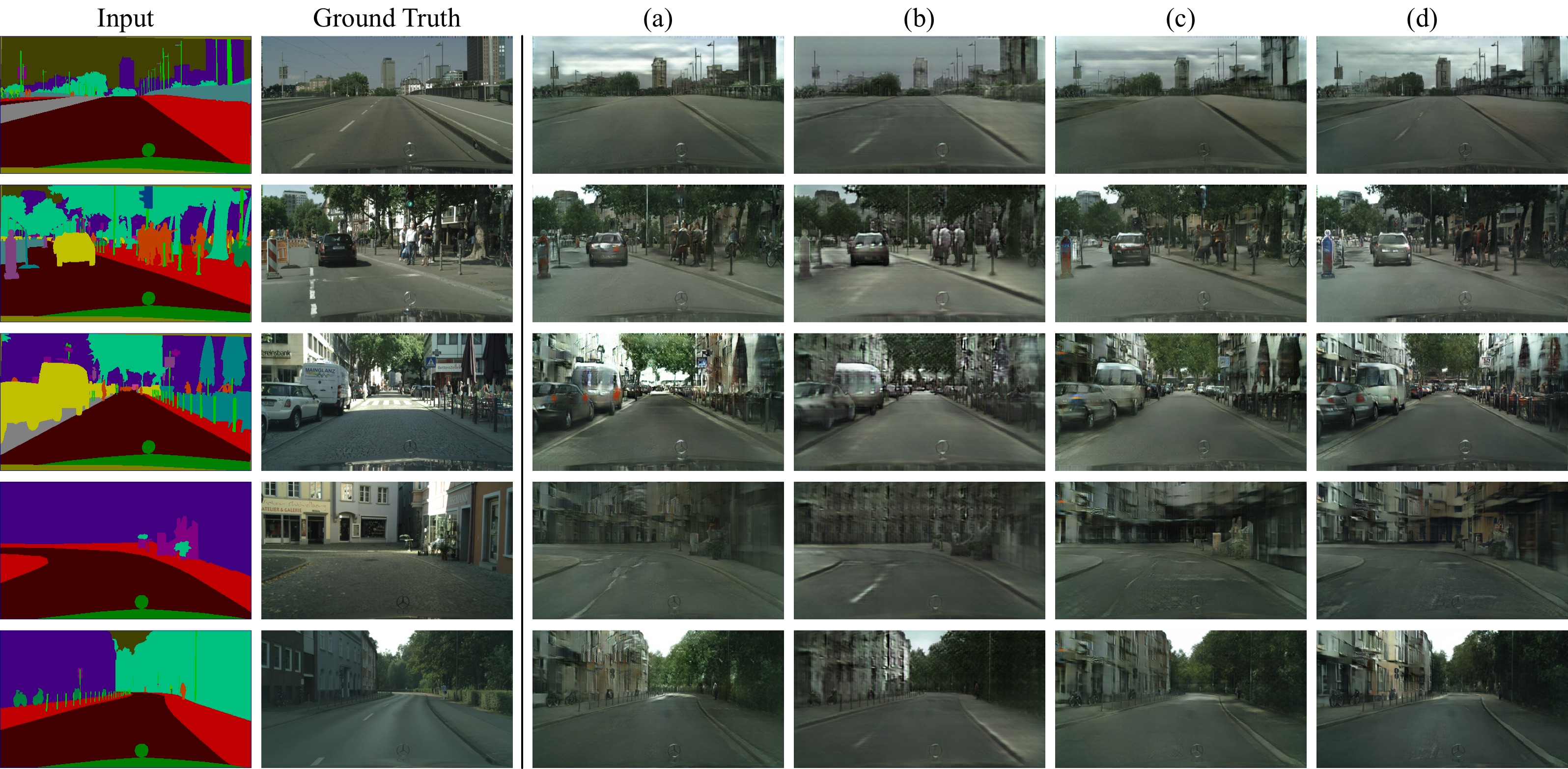}
\end{center}
\vspace{1mm}
\caption{
Visual examples synthesized by different methods with 10\% training data on the Cityscapes dataset. 
The columns of results are (a) SPADE Net~\cite{choi2018stargan} (baseline), (b) baseline + WM~(Equation \ref{eq:WM}), (c) baseline + mixup~\cite{zhang2017mixup}, and (d) baseline + ReMix (ours).
}
\label{fig:city1}
\end{figure*}